\pdfoutput=1
\documentclass[10.5pt]{article}

\usepackage{acl}
\usepackage{times}
\usepackage{latexsym}

\usepackage[T1]{fontenc}

\usepackage[utf8]{inputenc}
\usepackage{microtype}
\usepackage{listings}
\usepackage{fancyvrb}
\usepackage{graphicx}
\usepackage{multirow}
\usepackage{subcaption}
\usepackage{graphicx}
\usepackage{babel,blindtext}
\usepackage{enumitem}
\usepackage{colortbl}
\usepackage{caption}
\usepackage{hhline}
\usepackage{amsmath,amssymb,amsfonts}%
\usepackage{textcomp}%
\usepackage{url}%
\usepackage{stfloats}%
\usepackage{etoolbox}%
\usepackage{soul}%
\usepackage[utf8]{inputenc}%
\usepackage{microtype}%
\usepackage{fancyvrb}%
\usepackage{babel,blindtext}%
\usepackage{mathrsfs}%
\usepackage[title]{appendix}%
\usepackage{xcolor}%
\usepackage{textcomp}%
\usepackage{manyfoot}%
\usepackage{booktabs}%
\usepackage{algorithm}%
\usepackage{algorithmicx}%
\usepackage{algpseudocode}%
\usepackage{caption}
\usepackage{etoolbox}
\usepackage{soul}
\usepackage[para,online,flushleft]{threeparttable}
\usepackage{rotating} % Rotating table
\usepackage{anyfontsize}
\usepackage{graphicx}
\usepackage{subcaption} 
\usepackage{placeins}     
\usepackage{subcaption}
\usepackage{float}     
\usepackage{setspace}
\usepackage{ifthen}
\newboolean{PrintVersion}
\setboolean{PrintVersion}{false}
\usepackage{amsmath,amssymb,amstext} % Lots of math symbols and environments

\title{Beyond Accuracy: Interpreting Topic Representation in Suicide Ideation Detection Models}

\author{Hamideh Ghanadian{$^{1}$}, Isar Nejadgholi{$^{2}$},  Hussein Al Osman{$^3$} \\
  {$^{1,2,3}$}University of Ottawa, Ottawa, Canada\\
  {$^2$}National Research Council Canada, Ottawa, Canada \\
  \footnotesize\texttt{{$^{1,3}$}\{Hghan053, Hussein.alosman\}@uottawa.ca}\\   \footnotesize\texttt{{$^{2}$}isar.nejadgholi@nrc-cnrc.gc.ca}}

\begin{document}
\maketitle
\begin{abstract}
% Suicide ideation detection models are typically evaluated using aggregate performance metrics, yet little is known about how they internally represent psychologically meaningful risk factors. In high-stakes mental health applications, understanding internal representations is critical for safety, transparency, and responsible deployment. In this work, we move beyond accuracy and conduct a mechanistic analysis of suicide detection models trained on different type of datatset. We then analyze the geometry of models internal representation using UMAP visualizations and a cosine-based separability measure. Our results show that topic-aware augmentation increases the number, coherence, and directional separability of feature directions associated with underrepresented psychosocial risk factors such as immigration, family issues, and financial crisis. These findings suggest that augmentation not only improves performance of the models but also reshapes internal geometry toward clearer and more topic-selective representations. 
Suicide ideation detection models are typically evaluated using aggregate performance metrics, yet little is known about how they internally represent psychologically meaningful risk factors. In high-stakes mental health applications, understanding these internal representations is essential for safety, transparency, and responsible deployment. In this work, we move beyond accuracy and analyze how suicide detection models trained on original and topic-augmented datasets encode psychological risk factors in their internal representation space. Using visualization and geometric analysis, we examine the coherence and separability of topic-related features. Our results show that topic-aware augmentation increases the clarity and distinctness of underrepresented psychosocial risk factors such as immigration, family issues, and financial crisis. These findings suggest that augmentation not only improves model performance but also leads to more structured and interpretable internal representations.
\end{abstract}

\section{Introduction}\label{sec:intro}
% Despite the significant effort NLP research has put into facilitating mental health analysis, the best-performing models lack interpretability. In sensitive AI systems such as mental-health support tools, interpretability is necessary to enable debugging, safety assurance, and meaningful control over model behavior, particularly when harms arise from rare, context-dependent failure modes that are not detectable through aggregate performance metrics. Without interpretable models, developers cannot reliably identify spurious correlations, audit learned representations, or apply targeted safety interventions, undermining both accountability and responsible deployment \cite{rudin2019stop}.
Despite strong performance in mental health NLP, leading models remain largely uninterpretable. In high-stakes applications such as suicide ideation detection, interpretability is essential for identifying spurious correlations, auditing learned representations, and ensuring safe and accountable deployment \cite{rudin2019stop}.

Studies show that some Generative Large Language Models (GLLMs) such as GPT models have superior performance in classifying suicidality of the text data compared to other examined GLLMs \cite{ghanadian2023chatgpt}. However, they still fall short when compared to classifiers tailored for this task through supervised training. Therefore, in this research, we focus on investigating the interpretability of an existing fine-tuned classifiers.

Some research showed that existing datasets lack topic diversity, and some topics are over-represented in datasets while others are absent or underrepresented \cite{ghanadian2024socially}. They demonstrated that classifiers may develop a false causal relationship between an over-represented concept and a given label, leading to excessive dependence on that concept and potentially compromising classification accuracy \cite{nejadgholi2023concept}. Also, some classifiers might become under-sensitive to topics that are not well-represented in training sets \cite{nejadgholi2022towards}. Therefore, it is crucial to develop methods that can facilitate the study of how topics are formed and used in trained classifiers.

In suicidal ideation detection, certain risk factors such as depression, anxiety, relationship problems, and hopelessness are strong predictors of suicidal ideation and appear frequently in training corpora \cite{ghanadian2024socially}. These concepts are often over-represented in the suicidal class, making them susceptible to being learned as sufficient causes for suicidality\cite{ghanadian2025improving}. Consequently, the classifier may over-rely on these concepts and ignore the broader context, leading to reduced generalizability. Conversely, topics that appear rarely in the dataset such as racism and immigration may be under-represented in the model’s internal features. Therefore, it is necessary to understand how under-representation of topics translates into feature coherence in the latent space. This insight is necessary for diagnosing gaps in topic coverage. These observations suggest that suicidal-ideation classifiers may not always form clear internal representations of psychological risk factors. Some concepts might appear as coherent and separable directions, while others may remain entangled with unrelated features. 
% The structure of these representations matters for safety, transparency, and targeted interventions. Topic-aware augmentation may also influence how these concepts are encoded.

% These considerations motivate a closer examination of the geometry of the model’s latent space. We outline the theoretical foundations for understanding internal features, introduce dictionary learning as a tool for revealing them, and analyze how augmentation changes their clarity and separability.

Our two research questions regarding the interpretability of the suicidal detection models are as follows:
\begin{itemize}[leftmargin=*]
\item\textbf{{RQ1: How are psychological risk factors of suicidal ideation represented within the internal activation space of language models trained for suicide ideation detection?}}
\item\textbf{{RQ2: Does topic-aware synthetic data augmentation improve the clarity, separability, and specialization of risk-factor representations in the model’s internal space?}}
\end{itemize}

The main contributions of this study are as
follows:

\begin{itemize}[leftmargin=*]

\item We provide a mechanistic analysis of suicide detection models by examining their internal activation space rather than relying only on performance metrics. Using overcomplete sparse autoencoders, we extract interpretable feature directions and analyze how psychological risk factors are encoded in the model’s latent geometry.

\item We investigate feature specialization and monosemanticity in learned sparse representations. Through analysis of top-activating samples and UMAP visualizations, we show that certain features align with specific psychosocial topics and form coherent clusters in representation space.

\item We introduce a geometric separability framework to measure how clearly psychological concepts are represented. Using cosine distance between concept and non-concept means, we quantify representational clarity and demonstrate how topic-aware augmentation reshapes internal feature structure.

\end{itemize}

\section{Background and Related Work}\label{sec:background}

% Mechanistic interpretability seeks not only to correlate inputs with outputs, but to explain a model’s internal computation by identifying which intermediate structures are represented, how they compose, and how they causally influence behavior. In this line of work, analysis focuses on the geometry of a model’s internal representations.
A common working assumption in mechanistic interpretability is the linear representation hypothesis, which states that meaningful features often approximate linear directions in a model’s activation space. Here, the activation space refers to the vector space formed by the model’s internal activations at a specific layer, where each vector corresponds to the model’s internal state for a given token or input. Under this view, probing the geometry of activations can reveal human-describable concepts that the model uses internally \cite{olah2020zoom}.

Modern transformer models achieve strong performance by representing many latent features within fixed-width activation spaces. Empirical evidence suggests that models encode far more features than there are embedding dimensions \cite{elhage2023monosemanticity}. 
% An embedding is a specific type of activation, typically referring to input token representations used for downstream tasks; embeddings therefore live within activation space, but do not exhaust it. 
The apparent mismatch between the number of useful features and the dimensionality of the activation space motivates the superposition hypothesis. The superposition hypothesis states high-dimensional spaces contain many nearly orthogonal directions, allowing networks to store multiple features in overlapping subspaces\cite{bricken2023monosemantic}. When the number of useful features exceeds the available dimensions, features become entangled, producing polysemantic representations that respond to multiple concepts \cite{elhage2022toy}.

In this context, a feature is defined as a direction in activation space that corresponds to a recurring internal pattern used by the model. A feature is called monosemantic if it consistently corresponds to a single interpretable concept across inputs; otherwise, it is polysemantic. Polysemanticity arises naturally under superposition and can obscure the internal logic behind model predictions. For high-stakes applications such as suicide ideation detection, understanding whether psychological risk factors are represented in monosemantic or polysemantic forms is crucial for safety, transparency, and targeted intervention \cite{zhang2024beyond}.

% Dictionary learning via sparse autoencoders (SAEs) provides a principled method for exposing these underlying feature directions. SAEs aim to recover a set of basis directions whose sparse linear combination reconstructs the model’s internal activations. Each basis direction corresponds to a feature, while the scalar output of the encoder indicates the feature activation, that is, how strongly a given feature is present for a specific token or input. Because sparsity encourages only a small number of features to activate for any given input, dictionary learning is well suited for probing whether psychological risk factors such as anxiety or family issues emerge as identifiable and coherent geometric directions in activation space \cite{mudide2024efficient}.
Sparse autoencoders (SAEs) provide a principled way to uncover feature directions in a model’s activation space. They decompose internal activations into sparse combinations of learned basis directions, where each direction represents a feature and its activation indicates its presence in a given input. This sparsity makes SAEs well suited for identifying coherent psychological risk factors as geometric directions in representation space \cite{mudide2024efficient}.

Concept-based explanation methods aim to explain model behavior using human-aligned concepts rather than raw activations. Testing with Concept Activation Vectors (TCAV) 
% represents a foundational approach in this literature. TCAV 
defines a concept by gathering example instances, learns a linear separator in an internal activation space to obtain a concept direction, and then measures how sensitive a model’s prediction is to movement along that direction. Sensitivity is computed via directional derivatives and aggregated into a global score of concept importance for a class \cite{kim2018interpretability}.

In this paper, we adopt a core idea from concept-based explanation methods: defining a concept using example sets and contrasting it with a non-concept baseline. Rather than measuring a concept’s influence on final predictions, as in sensitivity-based approaches like TCAV, we assess whether the concept is geometrically well formed in the model’s internal representation. Sensitivity scores mix representational structure with decision-boundary effects and depend on task-specific gradients, limiting comparability across models and layers. In contrast, directional separability directly measures how distinctly a concept occupies activation space, independent of the classifier head. This makes separability a natural diagnostic for detecting entangled or underrepresented risk factors. 
% Moreover, it aligns with dictionary learning: a well-encoded concept should activate a coherent subset of feature directions whose average orientation diverges from unrelated text. This geometric view is particularly suited to evaluating how topic-aware augmentation reshapes internal representations.

\section{Datasets}\label{dataset}
\label{sec:datasets}

This work analyzes internal representations learned by models trained on a real-world suicide risk dataset and its augmented variant. 

\subsection{University of Maryland Reddit Suicidality Dataset (UMD)}

The University of Maryland Reddit Suicidality Dataset (UMD) \cite{zirikly2019clpsych,shing2018expert} is a collection of Reddit posts and comments authored by individuals expressing varying levels of suicide risk. The data are drawn primarily from mental health–related subreddits, including \textit{r/SuicideWatch}, and are annotated at the user level using a four-point ordinal scale: \textit{No Risk}, \textit{Low Risk}, \textit{Moderate Risk}, and \textit{High Risk}.

Following prior work\cite{ghanadian2023chatgpt}, we use the subset corresponding to Task A. In this setting, each user typically contributes a small number of posts, often limited to activity within \textit{r/SuicideWatch}. Because annotations are provided at the user level, all posts authored by a given user are concatenated into a single document, yielding one input instance per user.

This dataset serves as the base distribution for representation analysis in our study. One model is trained solely on UMD, while a second model is trained on an augmented version of the same data, enabling a controlled comparison of how data augmentation affects the geometry and sparsity of learned representations.

\subsection{Augmented Dataset}

The augmented dataset used in this study follows the topic-aware augmentation strategy introduced in \cite{ghanadian2024socially}. In that work, synthetic suicide-related posts were generated using large language models, with ChatGPT (zero-shot setting) producing the highest-quality synthetic dataset. The final augmented training set was constructed by combining this synthetic dataset with progressively increasing subsets of the original UMD training data. Specifically, 10\%, 20\%, and 30\% non-overlapping folds of UMD data were incrementally added to the synthetic corpus, and model performance was evaluated at each stage. The optimal balance was achieved at 30\% real data augmentation, where the model matched or surpassed the baseline trained on the full UMD dataset. In the present paper, we utilize augmented dataset presented in \cite{ghanadian2024socially} to  investigate how topic-aware augmentation reshapes internal feature representations.

\section{Methodology}\label{Methodology}
In this paper, we shift the focus from model outputs to the model’s internal representations, namely the intermediate activation vectors produced during processing. These activations live in the model’s residual-stream activation space and encode semantic and psychological structure implicitly. 
% Rather than treating suicidal ideation detection as a black-box classification problem, we analyze these internal representations to identify interpretable feature directions that correspond to recurring patterns in language use.

To expose such feature directions, we apply dictionary learning via sparse autoencoders to the model’s residual-stream activations. The sparse autoencoder decomposes each activation vector into a sparse linear combination of learned basis directions. Each basis direction, represented by a decoder vector, defines a feature in activation space, while the corresponding scalar output of the encoder indicates the feature activation for a given token. The resulting set of learned feature directions serves as the foundational object of analysis in the remainder of the methodology, enabling both qualitative inspection through top-activating text examples and quantitative geometric analysis.

Once these feature directions are learned, we employ two complementary methods to assess their structure and interpretability. First, we use Uniform Manifold Approximation and Projection (UMAP) \cite{mcinnes2018umap} as a non-linear dimensionality reduction technique to visualize relationships between feature directions. The input to UMAP consists of the normalized decoder vectors of the sparse autoencoder, so each point in the visualization corresponds to a single learned feature rather than to an input sample. 
% This visualization allows us to examine whether features associated with similar psychological topics occupy nearby regions of activation space, indicating more selective and structured representations, or whether they are dispersed, suggesting stronger superposition.
% UMAP is used here as an exploratory tool that emphasizes preservation of local neighborhood structure rather than faithful reconstruction of global geometry. As such, it 
UMAP provides an intuitive view of local clustering, overlap, and separation among feature directions that would be difficult to interpret directly in the original high-dimensional activation space.

Second, we move beyond visualization and introduce a quantitative analysis based on cosine distance. Inspired by Concept-based explanation methods, for each psychological concept, we construct two sets of input texts: a concept set containing samples labeled with that topic, and a non-concept set consisting of unrelated Reddit posts. Each input text is represented by an activation vector extracted from the same residual-stream layer of the model.
We compute an average representation for each set by taking the mean of the corresponding activation vectors, yielding one mean vector for the concept set and one for the non-concept set. These mean vectors live in the same residual-stream activation space and represent the average internal state associated with concept-related and unrelated content, respectively. We then compute the cosine distance between these two mean vectors, which measures their angular separation. Larger cosine distance indicates that the model encodes the concept in a direction that is more clearly separated from unrelated text.

\subsection{Dictionary Learning on Residual Streams} \label{sec: dictionary}

The linear representation hypothesis proposes that neural networks encode meaningful concepts as approximately linear directions in activation space. In this context, we refer to such directions as features. The superposition hypothesis further suggests that high-dimensional activation spaces contain many nearly orthogonal directions, enabling models to represent more features than the ambient dimensionality by packing them into overlapping subspaces. When many features compete for limited representational capacity, they may become entangled, leading to polysemantic representations.

These hypotheses motivate the use of dictionary learning to recover a set of basis directions such that internal activations can be reconstructed as sparse linear combinations of feature directions. The goal is not only to reconstruct activations accurately, but also to investigate whether the recovered directions correspond to interpretable and semantically meaningful patterns that help explain model behavior \cite{olshausen1997sparse}. Recent work has shown that this approach is effective for transformer models, where sparse autoencoders (SAEs) provide a scalable approximation for recovering interpretable and steerable feature directions from residual-stream activations \cite{bricken2023monosemantic, anthropic_scaling, openai2024sae}. In this study, sparse autoencoders serve as the core mechanism for dictionary learning and allow us to probe concept-like directions within the model’s residual-stream activation space.

\subsubsection{Sparse Autoencoder (SAE)}

Sparsity is a key constraint in sparse autoencoders. It enforces that only a small subset of latent units becomes active for any given input activation vector. This regularization prevents information from being diffusely distributed across many units and instead encourages each activation to be represented as a combination of a few distinct basis directions. 
% In practice, sparsity improves interpretability because each latent unit tends to correspond to a more localized and meaningful direction in activation space. 
% It also reduces interference between features and promotes disentanglement, which is critical for identifying semantically coherent feature directions. From an optimization perspective, sparsity acts as an information bottleneck that discourages redundant representations and leads to more stable and generalizable decompositions of internal activations.

% Overcompleteness refers to using a latent dimensionality that exceeds the dimensionality of the input activation space. 
In the context of transformer residual streams, overcompleteness is essential for addressing superposition, where many independent features compete for limited representational capacity. By expanding the number of latent units beyond the dimensionality of the residual stream, the autoencoder gains sufficient capacity to separate overlapping directions into more specific and independent features. Overcomplete sparse autoencoders therefore provide a principled mechanism for uncovering monosemantic feature directions that would otherwise remain entangled within dense model activations.

Let $x \in \mathbb{R}^D$ denote a model activation vector extracted from the residual stream, normalized so that the average squared $\ell_2$ norm equals the model dimension $D$. A sparse autoencoder learns $F$ latent features with $F \gg D$, yielding an overcomplete dictionary. The encoder maps the input activation vector to sparse feature activations via
\begin{equation}
f_i(x) ~=~ \mathrm{ReLU}\!\big( \langle W^{\mathrm{enc}}_{i,:}, x \rangle + b^{\mathrm{enc}}_i \big),
\label{eq:sae-encoder}
\end{equation}
where $W^{\mathrm{enc}} \in \mathbb{R}^{F \times D}$ and $b_{\mathrm{enc}} \in \mathbb{R}^F$ are learned parameters. The scalar quantity $f_i(x)$ represents the activation of feature $i$ for input $x$.

The decoder reconstructs the original activation vector as a linear combination of learned feature directions:
\begin{equation}
\hat{x} ~=~ b_{\mathrm{dec}} + \sum_{i=1}^{F} f_i(x)\, W^{\mathrm{dec}}_{:, i},
\label{eq:sae-decoder}
\end{equation}
where $W^{\mathrm{dec}} \in \mathbb{R}^{D \times F}$ and $b_{\mathrm{dec}} \in \mathbb{R}^D$. Each column $W^{\mathrm{dec}}_{:,i}$ defines a feature direction in residual-stream activation space.

The model is trained to minimize a loss function that combines reconstruction error with a sparsity penalty:
\begin{equation}
\mathcal{L} ~=~ \mathbb{E}_{x}\!\left[
\|x - \hat{x}\|_2^2
~+~
\lambda \sum_{i=1}^{F} f_i(x)\,\|W^{\mathrm{dec}}_{:,i}\|_2
\right],
\label{eq:sae-loss}
\end{equation}
where $\lambda$ controls the strength of the sparsity constraint.

Including $\|W^{\mathrm{dec}}_{:,i}\|_2$ inside the penalty allows the decoder vectors to be interpreted as normalized feature directions,
\begin{equation}
\tilde{W}^{\mathrm{dec}}_{:,i}
~=~
\frac{W^{\mathrm{dec}}_{:,i}}{\|W^{\mathrm{dec}}_{:,i}\|_2},
\label{eq:sae-normalized-decoder}
\end{equation}
with corresponding rescaled feature activations
\begin{equation}
a_i(x) ~=~ f_i(x)\,\|W^{\mathrm{dec}}_{:,i}\|_2.
\label{eq:sae-feature-activation}
\end{equation}

Under this decomposition, each token-level activation vector is explained by a small number of nonzero feature activations drawn from a large dictionary of feature directions. This parts-based representation aligns with the goal of identifying interpretable and potentially monosemantic directions within the model’s residual-stream activation space \cite{bricken2023monosemantic, anthropic_scaling}.

Overcompleteness ($F > D$) provides sufficient capacity to separate overlapping directions that arise under superposition, while sparsity ensures that only a small number of features activate for any given token. Empirically, increasing $F$ and appropriately tuning the sparsity coefficient improves feature specificity, reduces feature interference, and increases the effectiveness of causal interventions such as feature clamping \cite{anthropic_scaling, openai2024sae}.

\subsection{Feature--Topic Association} \label{sec: feature-topic}

To relate learned feature directions to psychological risk factors, we explicitly define a mapping between sparse autoencoder features and annotated topics. 
% This step is essential for interpreting the geometry of the learned feature space and for attributing semantic meaning to individual feature directions.

Let $i$ index a learned feature direction, and let $a_i(x)$ denote the rescaled activation of feature $i$ for input sample $x$ as defined in Equation~\ref{eq:sae-feature-activation}. Each sample $x$ is associated with a psychological topic label $t \in \mathcal{T}$, where $\mathcal{T}$ denotes the set of all annotated topics.

For each feature $i$ and each topic $t$, we compute the average feature activation over all samples belonging to that topic:
\begin{equation}
\bar{a}_{i,t} ~=~ \mathbb{E}_{x \sim \mathcal{D}_t}\!\left[ a_i(x) \right],
\end{equation}
where $\mathcal{D}_t$ denotes the set of samples labeled with topic $t$. This quantity measures how strongly feature $i$ responds, on average, to content associated with topic $t$.

We define the \emph{dominant topic} of feature $i$ as the topic for which this average activation is maximal:
\begin{equation}
t^*(i) ~=~ \arg\max_{t \in \mathcal{T}} \; \bar{a}_{i,t}.
\end{equation}

This assignment yields a many-to-one mapping from feature directions to psychological topics. Importantly, this mapping is descriptive rather than exclusive: a feature may respond to multiple topics, but the dominant topic captures the strongest association and provides a principled basis for visualization and analysis.

% This feature-topic association is used consistently throughout the remainder of the paper. In visualization, features are colored according to their dominant topic to reveal geometric structure in the learned representation space.

\subsection{UMAP for Latent Feature Visualization} \label{sec: umap}

While sparse autoencoders allow us to extract interpretable feature directions from the model’s residual stream activation space, these directions live in a high-dimensional space that is difficult to inspect directly. To gain intuition about the organization of the learned feature directions, we employ Uniform Manifold Approximation and Projection (UMAP), a non-linear dimensionality reduction technique designed to preserve local neighborhood structure while revealing coarse geometric patterns \cite{mcinnes2018umap}.

In this work, UMAP is applied to the set of learned feature directions, represented by the normalized decoder vectors $\tilde{W}^{\mathrm{dec}}_{:,i}$. Each feature direction is therefore treated as a point in residual-stream activation space, and UMAP provides a two-dimensional embedding that reflects similarities between feature directions. 
% We use cosine distance as the similarity metric, as angular relationships between directions are more meaningful than Euclidean distance in high-dimensional representation spaces and are consistent with the linear representation hypothesis.
UMAP serves a primarily exploratory and qualitative role in our methodology. It allows us to visually assess whether feature directions associated with similar psychological topics occupy nearby regions of activation space, indicating structured and selective representations, or whether they are dispersed, suggesting stronger superposition and polysemantic encoding. By comparing UMAP embeddings across models trained on different datasets, we can further examine how data augmentation affects the geometric organization of risk-related feature directions. Importantly, UMAP is not used to define or enforce feature structure, but rather to reveal patterns that emerge naturally from the learned representations.

\subsection{Cosine Distance for Quantitative Separability Analysis} \label{sec: cosine}

% While UMAP provides an intuitive visualization of relationships between feature directions, it does not offer a quantitative measure of how clearly different psychological concepts are separated in the model’s internal representation space. In residual-stream activation space and in sparse autoencoder–derived representations, vector norms can vary across tokens, samples, and datasets. Cosine distance removes the effect of magnitude and isolates whether a concept is encoded along a direction that differs from a non-concept baseline.

% Cosine distance is well suited to linear representation settings. If internal representations correspond to approximately linear directions, then the angle between mean representations summarizes how consistently those directions distinguish one set of samples from another. Larger cosine distance indicates that the two mean vectors are oriented further apart in activation space, implying that a simple linear decision boundary could separate them with a larger angular margin. This aligns with the goals of interpretability, as stable directional differences are easier to analyze and relate to specific feature directions recovered by the sparse autoencoder.
Cosine distance is well suited to linear representation settings because it measures angular differences between mean activation vectors. If internal representations align with approximately linear directions, larger cosine distance indicates clearer directional separation between concept and non-concept sets. This makes separability easier to interpret and directly relatable to feature directions recovered by the sparse autoencoder. UMAP provides an intuitive visualization of feature relationships but does not quantify how clearly psychological concepts are separated. Cosine distance addresses this by measuring directional differences while ignoring magnitude, allowing us to assess whether a concept is encoded distinctly from a non-concept baseline.
% To complement the qualitative insights from visualization, we introduce a quantitative analysis based on cosine distance between mean activation vectors. For each psychological concept, we define a concept set consisting of samples associated with that concept and a corresponding non-concept set drawn from unrelated content. Using the rescaled feature activations defined in Equation~\ref{eq:sae-encoder}, we aggregate activation vectors across samples to compute an average latent direction for each set. The cosine distance between the resulting concept and non-concept mean vectors provides a measure of angular separability in residual-stream activation space.

To complement the qualitative insights from visualization, we introduce a quantitative analysis based on cosine distance computed in the sparse autoencoder (SAE) feature space. For each psychological concept, we define a concept set consisting of posts associated with that concept and a corresponding non-concept set drawn from unrelated Reddit content.
For each post, token-level residual-stream activations are first mapped into the SAE latent space, yielding a vector of feature activations. 

Representations are extracted from the same model layer for both concept and non-concept sets and mapped into the sparse autoencoder feature space. Within this space, we compute mean feature activation vectors for the concept set and the non-concept set. To obtain post-level feature activation representations, we aggregate token-level feature activations within each post using a max-pooling operation. For each feature, the post-level activation is defined as the maximum activation across all tokens in the post. This choice reflects the assumption that the presence of a psychologically meaningful feature anywhere in a post is sufficient to indicate its relevance and avoids dilution of sparse but salient signals that would otherwise occur with averaging across tokens.

For each psychological topic, we then compute a mean post-level feature activation vector by averaging these post-level representations across all posts in the concept set, yielding $\mu_{\text{concept}}$. An analogous mean feature activation vector $\mu_{\text{non}}$ is computed from the non-concept baseline. Larger cosine distances indicate that concept-related and non-concept content activate more distinct sets of interpretable features, reflecting clearer separation in the model’s internal representation.
Separation between the two is quantified using cosine distance,
\begin{equation}
d_{\cos} \;=\; 1 - \frac{\mu_{\text{concept}} \cdot \mu_{\text{non}}}{\lVert \mu_{\text{concept}} \rVert_2 \, \lVert \mu_{\text{non}} \rVert_2}.
\end{equation}

% This construction follows the core idea introduced in TCAV, where concepts are defined using example sets and represented by averaged directions in activation space; however, 
Despite TCAV methods that measure sensitivity of model outputs to these directions, we use cosine distance to quantify their geometric separability within the model’s internal activation space. Prior work in representation analysis and concept-based interpretability has similarly relied on cosine similarity or distance to assess alignment between learned representations and semantic concepts \cite{kim2018interpretability, bricken2023monosemantic}.

Larger values indicate greater separation between the concept and the non-concept baseline. We compute $d_{\cos}$ for both the UMD fine-tuned model and the AUG fine-tuned model, each evaluated on the same mixed synthetic test set, and repeat this procedure for every topic to obtain a cosine distance per topic.

\subsection{Experimental Setup}\label{sec: data}

In our Experimentation we utilize ALBERT model(Albert-base-V2)\footnote{\href{https://huggingface.co/albert/albert-base-v2}{AlBERT Model}}, a lightweight transformer model that reduces parameter redundancy through cross-layer parameter sharing and factorized embedding parameterization. We fine-tune ALBERT separately on the UMD dataset and the augmented dataset \cite{ghanadian2025improving}. ALBERT is particularly suitable for our interpretability analysis because its compact architecture and shared layers produce stable residual-stream representations, enabling clearer examination of learned feature directions through sparse autoencoders.

We train a sparse autoencoder on token-level residual-stream activations from a fine-tuned ALBERT classifier in order to recover an overcomplete set of feature directions that can be inspected and analyzed. 
% ALBERT is fine-tuned on the UMD suicidal ideation dataset and Augmented dataset, after which token-level residual-stream activations are extracted from the final transformer layer.
We focus on the final-layer residual stream because it captures high-level semantic representations that are most relevant to the downstream suicide ideation detection task, while remaining upstream of the classification head. This choice allows us to analyze psychologically meaningful directions that the model uses for decision-making, without conflating them with task-specific output weights. Prior work in mechanistic interpretability has similarly shown that late-layer residual streams contain rich, interpretable features aligned with abstract concepts, making them suitable targets for sparse dictionary learning \cite{bricken2023monosemantic, templeton2024scaling}.

We adopt the encoder–decoder architecture described in Section~\ref{sec: dictionary}. An overcomplete feature dictionary is used, with $F = 4096$ learned feature directions for input activation vectors of dimension $D = 768$. This expansion increases the representational capacity of the dictionary and improves the separation of overlapping directions that arise due to superposition, allowing more specific and interpretable feature directions to emerge. The token dimension is flattened and filtered using the attention mask so that only non-padding token activations are fed to the sparse autoencoder. The encoder consists of a linear map followed by a ReLU nonlinearity, yielding non-negative and selective feature activations. The decoder is a linear map that reconstructs the original activation vector from a sparse linear combination of feature directions. Under this formulation, the decoder columns represent candidate feature directions, while the encoder outputs indicate how strongly each feature direction is activated for a given token.

Optimization is performed using the AdamW optimizer, which provides stable convergence and improved control over model capacity by decoupling weight decay from gradient updates. Early stopping is applied based on validation reconstruction loss to prevent overfitting once learning plateaus. After training, feature activations are computed for all tokens in the dataset and aggregated at the sample (post) level by taking the maximum activation per feature across tokens within each text. This aggregation reflects the assumption that a feature is present in a sample if it activates strongly for at least one token.

For each learned feature direction, we then select the top 100 samples with the highest feature activations. This procedure follows established practice in sparse autoencoder interpretability work, where a fixed number of top-activating examples has been shown to be sufficient for reliably characterizing feature semantics while remaining computationally tractable \cite{templeton2024scaling, bricken2023monosemantic}.

\section{Results}
\subsection{Analysis I: Latent Feature Visualization}

A central question in this analysis is whether suicide-related psychological risk factors such as hopelessness, anxiety, and social isolation form coherent geometric structure in the model’s learned feature directions. We also ask whether a topic-aware data augmentation improves the coherence of these features.

For each model—ALBERT fine-tuned on the UMD dataset and ALBERT fine-tuned on the augmented dataset—we assign a dominant psychological topic to each learned feature direction using the following procedure. First, for a given feature, we collect its activation values across all posts in the dataset. Second, posts are grouped according to their annotated psychological topic. For each topic, we compute the average feature activation over all posts belonging to that topic. The dominant topic of the feature is defined as the topic for which this average activation is highest. This assignment provides a coarse but interpretable mapping between learned feature directions and psychological risk factors.

Throughout this paper, sparse autoencoders are trained on token-level residual-stream activations in order to capture fine-grained linguistic features. However, psychological topics are defined at the post level rather than at the token level. To bridge this gap, token-level feature activations are aggregated within each post using a max-pooling operation, yielding a post-level feature activation vector. All topic assignments, visualizations, and quantitative analyses reported in this paper are therefore performed at the post level, unless explicitly stated otherwise.

\begin{figure*}[ht]
  \centering
  \caption{UMAP projections of features highlighting three selected topics. Left column: ALBERT fine-tuned on AUG (the augmented dataset with high topic coverage). Right column: ALBERT fine-tuned on UMD. Colored points mark features whose dominant responses align with the indicated topic; dark blue points show all other features.}
  \label{fig: feature_vis}
  \begin{subfigure}[t]{0.32\linewidth}
    \centering \includegraphics[width=\linewidth,height=0.28\textheight,keepaspectratio]{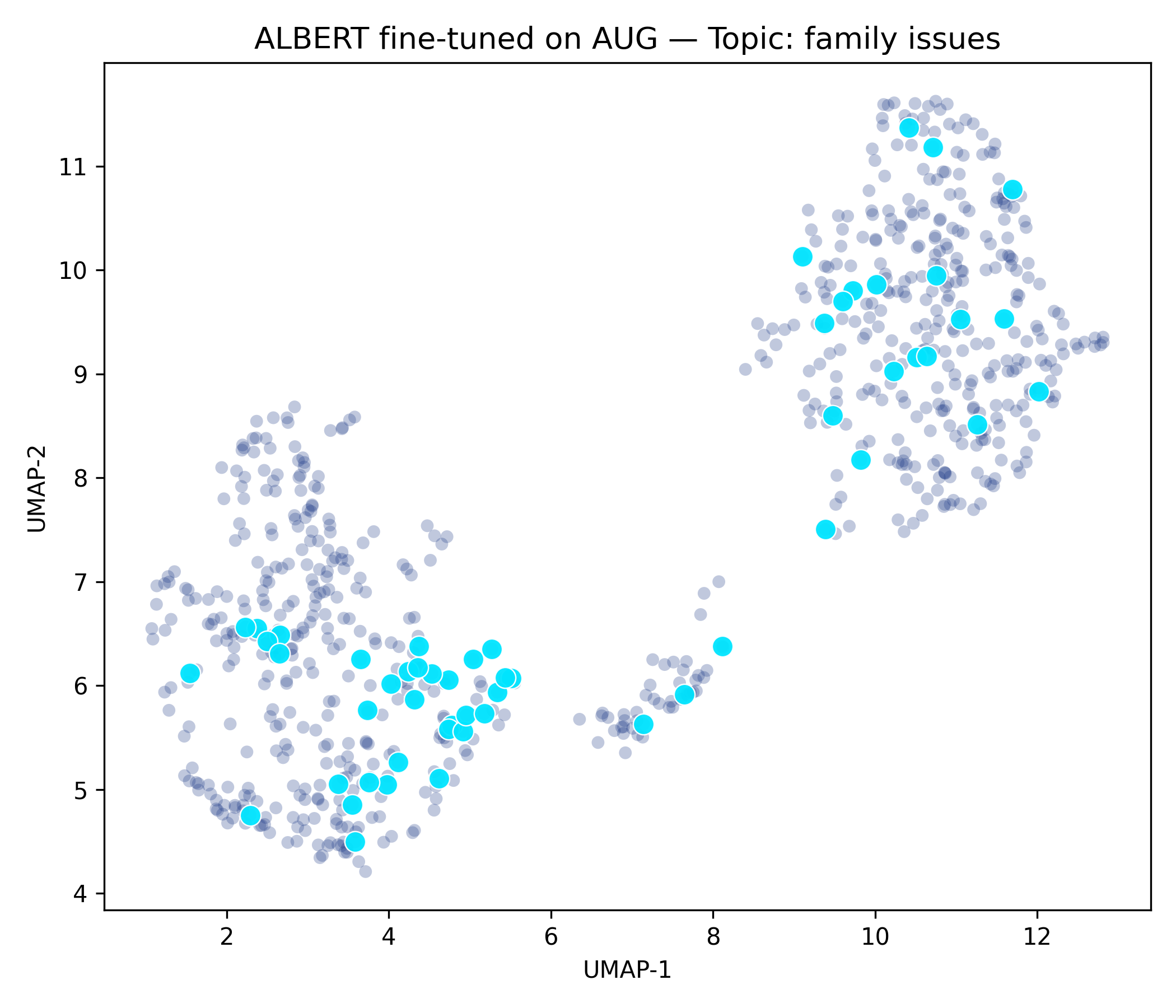}
  \end{subfigure}\hfill
  \begin{subfigure}[t]{0.32\linewidth}
    \centering \includegraphics[width=\linewidth,height=0.28\textheight,keepaspectratio]{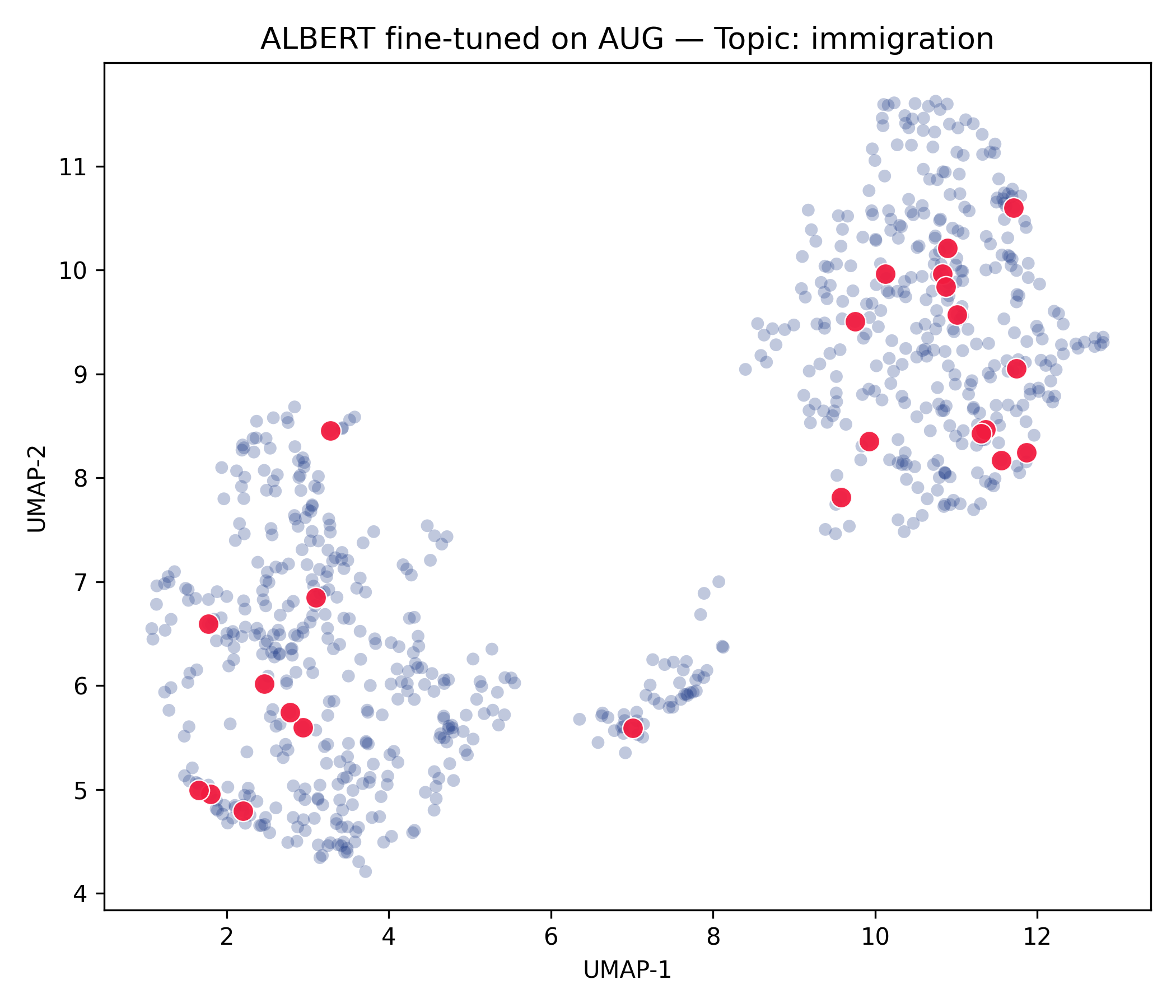}
  \end{subfigure}\hfill
   \begin{subfigure}[t]{0.32\linewidth}
    \centering \includegraphics[width=\linewidth,height=0.28\textheight,keepaspectratio]{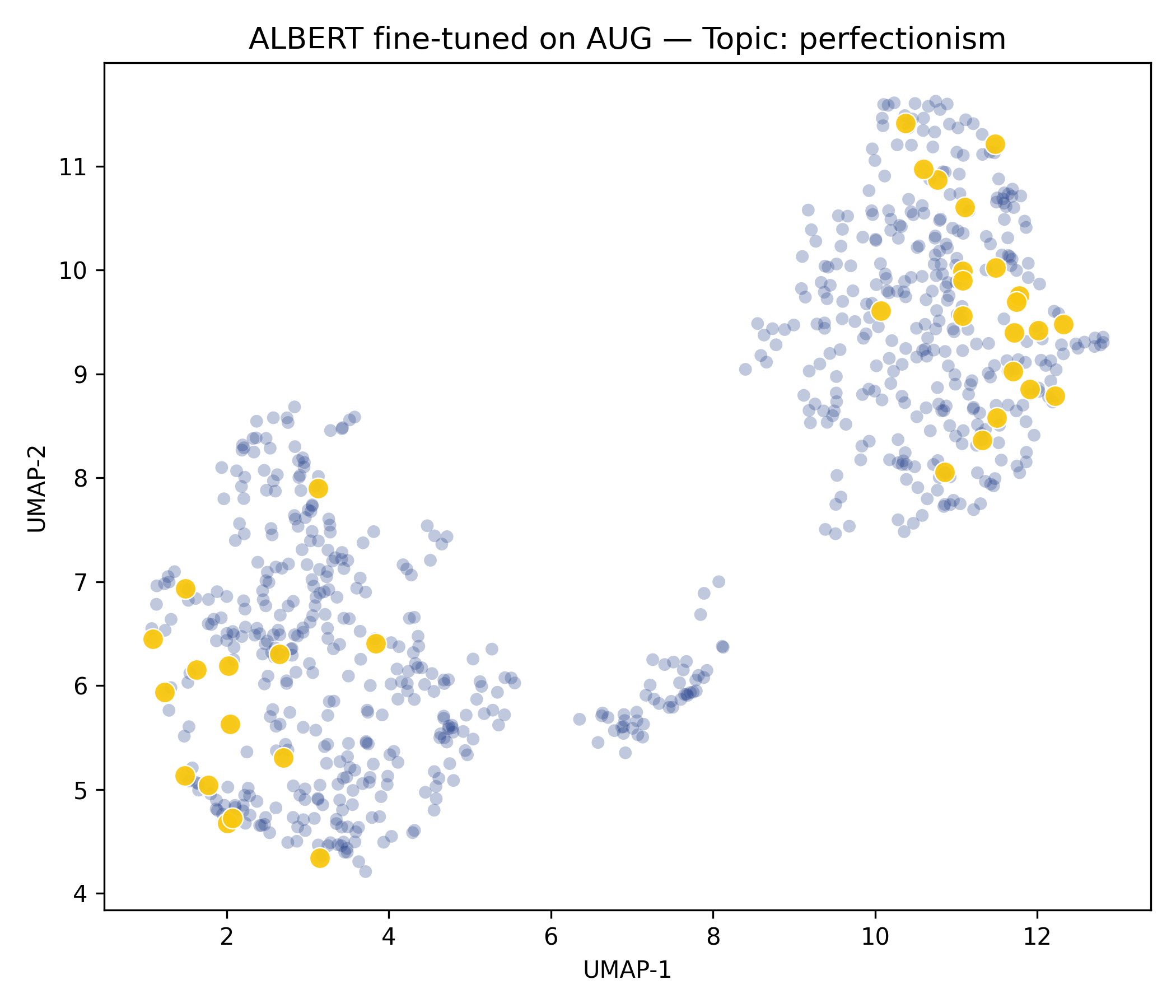}
  \end{subfigure}\hfill
    \begin{subfigure}[t]{0.32\linewidth}
    \centering \includegraphics[width=\linewidth,height=0.28\textheight,keepaspectratio]{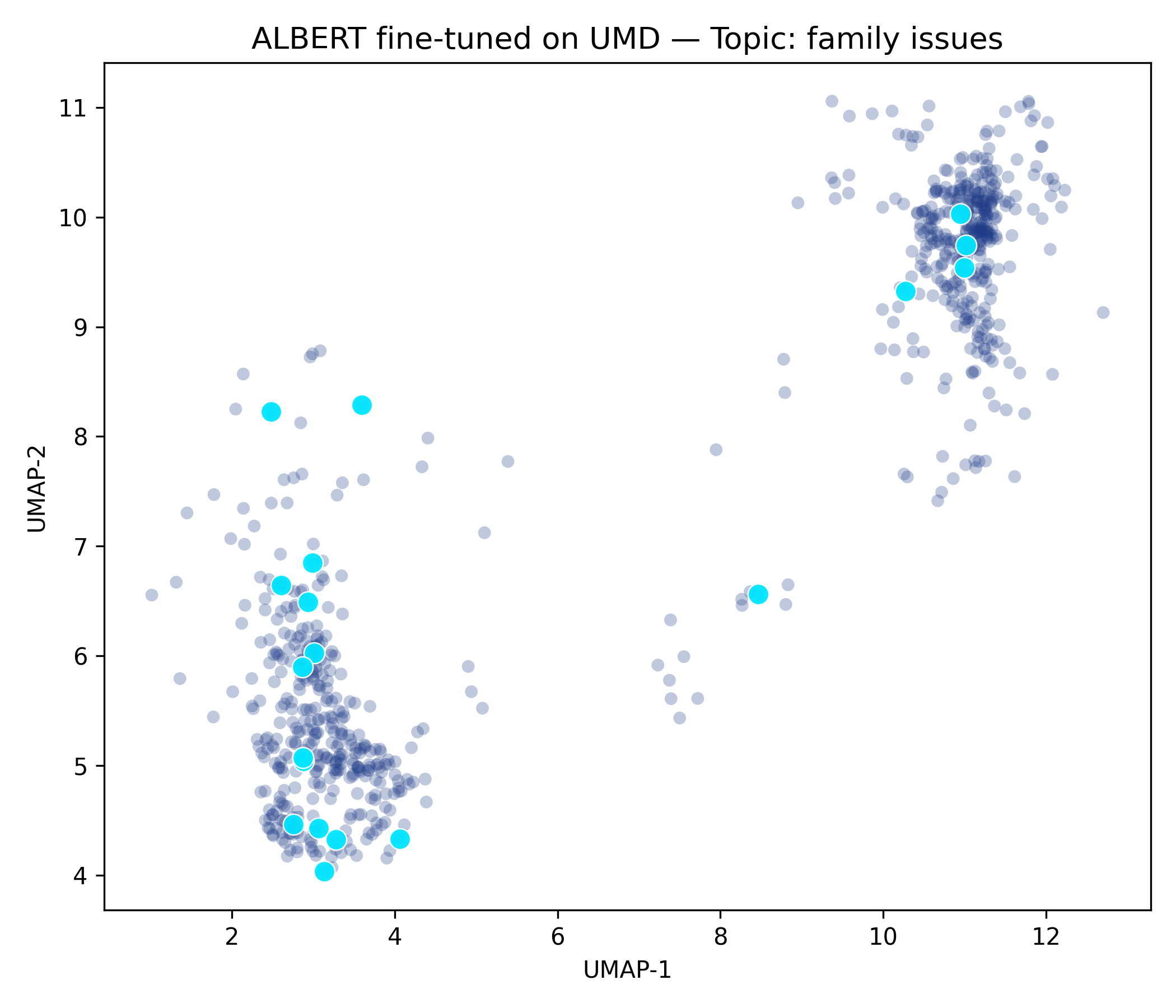}
  \end{subfigure}\hfill 
  \begin{subfigure}[t]{0.32\linewidth}
    \centering \includegraphics[width=\linewidth,height=0.28\textheight,keepaspectratio]{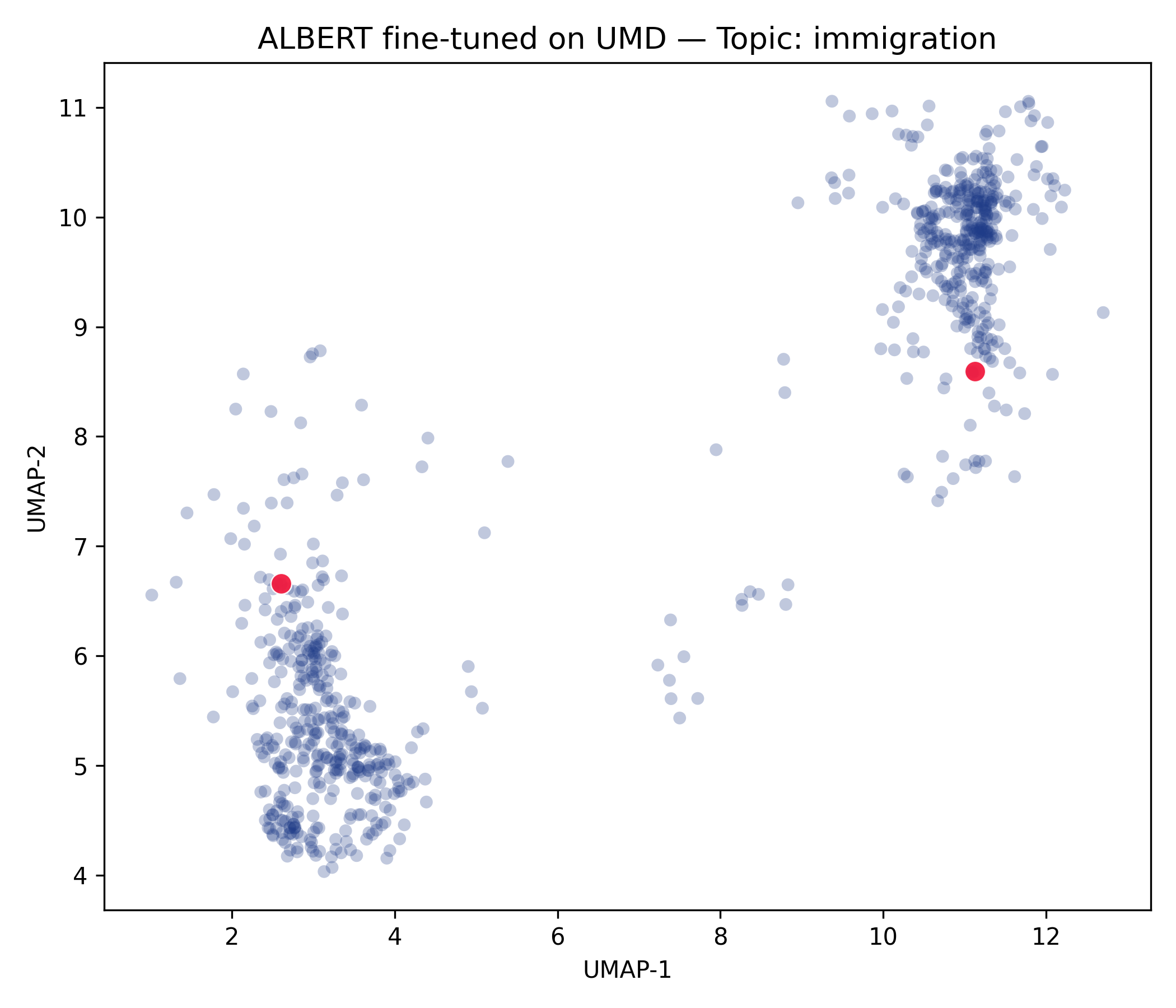}
  \end{subfigure}\hfill
  \begin{subfigure}[t]{0.32\linewidth}
    \centering \includegraphics[width=\linewidth,height=0.28\textheight,keepaspectratio]{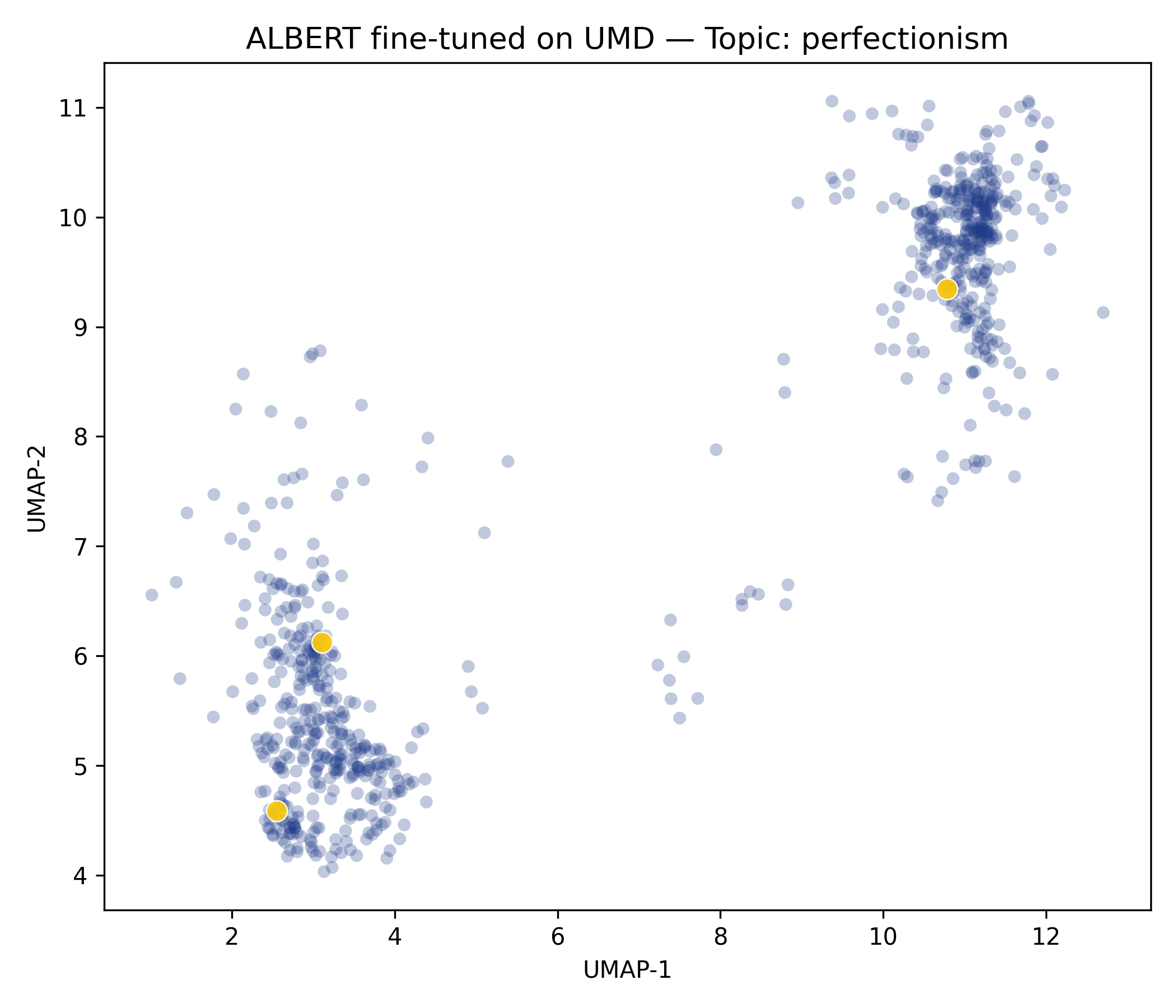}
  \end{subfigure}
\end{figure*}

After assigning dominant topics, the full set of feature directions (decoder vectors) is projected into two dimensions using UMAP. In Figure~\ref{fig: feature_vis}, we visualize the geometry of the feature space one topic at a time by highlighting features whose dominant topic matches the topic of interest, while rendering all remaining feature directions as a low-opacity background. Compact highlighted regions indicate feature directions that respond selectively to the same psychological concept, suggesting movement toward monosemantic structure. In contrast, dispersed highlighted features suggest shared or overlapping representations, consistent with superposition and polysemantic encoding. For each topic, we present paired UMAP panels, with the AUG-trained model shown on the left and the UMD-trained model on the right, enabling direct comparison of how data augmentation affects the geometric organization of risk-related feature directions.

In Figure~\ref{fig: feature_vis}, each point corresponds to a learned feature direction, represented by its decoder vector in the model’s residual-stream space. The colored points indicate features whose dominant activations are associated with the indicated psychological topic. In the UMD fine-tuned model, topics such as perfectionism and immigration activate only a small number of scattered feature directions, suggesting limited topic coverage and stronger superposition, as the same directions appear to respond to multiple phenomena. 

In contrast, the augmented fine-tuned model exhibits a larger number of highlighted feature directions for the same topics, which often form compact regions within the two macro-clusters of the embedding. These regions indicate sets of directions that respond more selectively to the target topic. The family issues topic follows a similar pattern, with sparse and dispersed feature directions in the UMD model and denser, more coherent groupings in the AUG model. This shift suggests that data augmentation improves both the availability and the consistency of topic-specific directions in the learned representation space.

Monosemanticity is valuable because a compact, well-separated group of topic-selective features supports clear explanations, stable linear steering, and predictable causal edits. However, polysemanticity is not inherently undesirable. Distributed highlights can reflect useful sharing of linguistic substrates, overlapping risk factors, or genuine subtopics that co-occur in real text. Such sharing can improve generalization and parameter efficiency when concepts partially overlap. The practical concern is interference rather than polysemanticity itself; if the same features fire for unrelated content, the representation becomes noisy and separation weakens. The visualizations indicate that augmentation moved several topics toward monosemanticity by increasing the number and coherence of topic-selective directions, while still preserving beneficial sharing where concepts naturally overlap.

\subsection{Analysis II: Quantifying Concept Separability}

The goal of this section is to quantify how clearly each psychological topic is separated from generic, non-risk-related text in the model’s internal activation space. We use cosine distance as a measure of separability because it captures directional differences between mean activation vectors associated with concept-related and non-concept-related content. If the mean vectors corresponding to two sets of points in noticeably different directions, this indicates that the model encodes them as distinct internal representations. Because cosine distance depends only on direction, it is robust to changes in vector magnitude caused by scaling, preprocessing, or dataset-specific norm shifts.

For post-level analysis, token-level activation vectors within each post are averaged to obtain a single activation vector per post. We evaluate two models that are fine-tuned separately: one model fine-tuned on the UMD dataset and one model fine-tuned on the augmented dataset. All measurements are computed on a held-out test set consisting of a mixture of synthetic posts annotated with psychological topic labels. No training or validation data are used in these analyses.

In addition to the synthetic dataset with topic labels, we use a separate Reddit dataset, which contains content unrelated to suicidal ideation. This Reddit set is treated as the non-concept baseline and is collected from unrelated subreddits to suicidal ideation. We merge the synthetic topic-labeled posts and the Reddit posts to form the final test set used for analysis. For each psychological topic, we define the \textbf{concept set} as all test posts labeled with that topic, and the \textbf{non-concept set} as the Reddit test set.

% \begin{figure*}[ht]
% \caption{Topic separation on the synthetic test set. Left and middle panels show cosine distance between each topic’s mean activation vector and a non-concept Reddit baseline for the UMD- and AUG-fine-tuned models (shared scale; higher indicates greater separation). The right panel shows the change $\Delta=\text{AUG}-\text{UMD}$ (diverging scale centered at 0), with topics sorted by $\Delta$.}
% \label{fig: heatmap}
% \centering
% {\includegraphics[width=0.8\linewidth]{Figures/heatmap.png}}
% \captionsetup{font=small}
% \end{figure*}

\begin{table}[h]
\caption{Cosine distance between topic means and a general Reddit baseline for models fine-tuned on AUG and UMD, evaluated on the synthetic test set. Distances reflect angular separation in representation space (higher is better). $\Delta$ is AUG–UMD and summarizes the change due to augmentation.}
\resizebox{\columnwidth}{!}{%
\label{tab: Cosine distance}
\centering
\begin{tabular}{c|ccc}
Topic & \begin{tabular}[c]{@{}c@{}}Cosine Distance\\  in AUG\end{tabular} & \begin{tabular}[c]{@{}c@{}}Cosine Distance\\  in UMD\end{tabular} & $\Delta$(AUG–UMD) \\ \hline
Relationship Problems & 0.513 & 0.178 & 0.334 \\ \hline
Financial Crisis & 0.524 & 0.213 & 0.310 \\ \hline
Family Issues & 0.466 & 0.199 & 0.266 \\ \hline
Anxiety & 0.268 & 0.0492 & 0.218 \\ \hline
Racism & 0.239 & 0.024 & 0.215 \\ \hline
Anger & 0.237 & 0.033 & 0.204 \\ \hline
Bullying & 0.206 & 0.019 & 0.186 \\ \hline
Immigration & 0.199 & 0.015 & 0.184 \\ \hline
Perfectionism & 0.153 & 0.041 & 0.113 \\ \hline
Death of Closed One & 0.212 & 0.108 & 0.105 \\ \hline
Education & 0.125 & 0.077 & 0.048 \\ \hline
Hopelessness & 0.200 & 0.217 & -0.017 \\ \hline
Unemployment & 0.824 & 0.903 & -0.079 \\ \hline
\end{tabular}
}
\end{table}

% Figure~\ref{fig: heatmap} summarizes topic–non-concept separation on the synthetic test set by showing the cosine distance between each topic’s mean activation vector and a non-concept Reddit baseline. 
Table~\ref{tab: Cosine distance} summarizes topic separation on the synthetic test set by showing the cosine distance between each topic’s mean SAE feature activation vector and a non-concept Reddit baseline.
Higher cosine distance indicates stronger directional separation of a topic from general Reddit content, reflecting more selective internal representations. 
% The left and middle panels report these distances for models fine-tuned on the original UMD data and on the augmented (AUG) data, respectively, using a shared color scale. The right panel visualizes the change induced by augmentation, defined as $\Delta = \text{AUG} - \text{UMD}$, with topics sorted by their change in separation.

Across most topics, augmentation leads to a clear increase in directional separation. The largest gains are observed for relationship problems ($+0.334$), financial crisis ($+0.311$), and family issues ($+0.266$), indicating that augmentation substantially sharpens the representation of these psychosocial factors relative to general Reddit content. Additional consistent improvements are visible for anxiety, racism, anger, bullying, and immigration (approximately $+0.18$ to $+0.22$), suggesting that topic-aware augmentation broadly improves the model’s ability to isolate diverse risk-related themes in its internal geometry.

More moderate gains are observed for perfectionism ($+0.113$), death of a close one ($+0.105$), depression ($+0.092$), and education ($+0.048$). These smaller increases indicate partial improvement in separability, consistent with topics that are either less lexically distinctive or more diffusely expressed across posts.

Two topics deviate from the overall trend. Hopelessness shows a slight decrease ($-0.017$), likely reflecting its strong semantic overlap with other risk factors such as depression and anxiety, which limits how distinctly it can be separated as an independent direction. Unemployment exhibits a larger decrease ($-0.079$); however, this topic already displays very high separation in both models (approximately $0.82$–$0.90$). Because its baseline separation is near the upper bound of the scale, there is limited room for further improvement, and small fluctuations appear as a decrease despite the topic remaining strongly isolated from the non-concept baseline.

\section{Discussions and Conclusion}\label{sec:Discussion}

This paper examined how suicide ideation detection models encode psychologically meaningful topics and how these representations change under topic-aware augmentation. Grounded in the linear representation and superposition hypotheses, we used overcomplete sparse autoencoders to recover feature directions from residual activations and analyzed their structure using UMAP visualizations and cosine-based separability. Together, these methods allowed us to evaluate not only which psychological topics are encoded in the model’s latent space, but also how selectively and distinctly they are represented.

Three findings emerge. First, UMAP visualizations show that topic-aware augmentation increases the number and compactness of topic-selective feature clusters, indicating a shift toward more structured and monosemantic representations. Second, cosine analysis confirms this effect quantitatively: augmentation generally increases directional separation between topic means and a general Reddit baseline, with the largest gains observed for relationship problems, financial crisis, and family issues. Third, some topics behave differently. Hopelessness shows little change, likely due to semantic overlap with other risk factors, while unemployment decreases slightly from an already high baseline, suggesting a ceiling effect rather than representational degradation.

These results refine our understanding of interpretability. While monosemantic features support explanation and intervention, some polysemantic sharing may reflect natural topic overlap and aid generalization. The key concern is not superposition itself, but harmful interference that blurs concept boundaries. Our findings suggest that topic-aware augmentation reduces such interference for many risk factors while preserving meaningful overlap, thereby improving both coverage and separability of psychological topics in the model’s internal geometry without imposing overly rigid representations.
\section{Future Works}

This paper explored suicide ideation detection from multiple complementary perspectives, including the use of large language models for classification, synthetic data generation to address dataset limitations, and representation-level analysis to better understand learned decision signals. While the results demonstrate the effectiveness of topic-aware augmentation and highlight the potential of both generative and discriminative models, several important directions remain for future research.

This paper provides insight into how topic-aware augmentation influences internal representations. Future work could build on this analysis by introducing causal interventions on learned features, such as clamping or amplifying sparse autoencoder activations, to assess their direct impact on model predictions. This would strengthen the connection between descriptive interpretability and functional model behavior, helping distinguish features that are merely correlated with suicidality from those that actively influence decisions.

The analyses in this paper focused on individual posts treated as independent samples. However, suicidal ideation often manifests through evolving language patterns over time. Future research could extend both classification and representation analyses to longitudinal user-level data, examining how risk-related signals accumulate or change across sequences of posts. This may enable earlier detection of emerging risk and provide a more realistic modeling framework for real-world applications.

Finally, future work should examine how both classifiers and synthetic-data-driven models perform across different populations, platforms, and linguistic styles. Assessing robustness to domain shifts and potential demographic biases is particularly important in mental health applications. Incorporating fairness-aware evaluation alongside performance and interpretability analyses would help ensure that improvements in accuracy do not come at the expense of equitable and responsible deployment.

\section{Limitations}

Despite providing insight into internal representations, the interpretability analysis in this thesis has several limitations. First, sparse autoencoders approximate the model’s internal structure but do not guarantee recovery of true underlying computational features. Learned features may still reflect mixtures of signals or residual polysemanticity. Second, cosine-based separability measures capture directional differences but do not establish causal importance for model predictions. Third, UMAP visualizations, while useful for understanding structure, depend on hyperparameters and dimensionality reduction choices that may influence observed cluster patterns. Moreover, the analysis was conducted on a specific model architecture and dataset configuration, and therefore the findings may not fully generalize to other architectures, scales, or domains.

\nocite{*} 
\bibliography{Ref}
\bibliographystyle{acl_natbib}

\end{document}